\def\year{2020}\relax
\documentclass[letterpaper]{article} 
\usepackage{aaai20}  
\usepackage{times}  
\usepackage{helvet} 
\usepackage{courier}  
\usepackage[hyphens]{url}  
\usepackage{graphicx} 
\usepackage{graphicx}  
\usepackage{subfigure}
\usepackage{amsmath}
\usepackage{amssymb}
\usepackage{amsthm}
\usepackage{algorithm}
\usepackage{algorithmic}
\usepackage{dsfont}

\theoremstyle{definition}
\newtheorem*{definition}{Definition}

\usepackage{url}
\usepackage{adjustbox}
\usepackage{multirow}

\title{Integrating Deep Learning with Logic Fusion for Information Extraction}
\author{Wenya Wang
        {\normalfont and} Sinno Jialin Pan
        \\
        Nanyang Technological University, Singapore
        \\
        \{wangwy, sinnopan\}@ntu.edu.sg
}
\begin{document}

\maketitle

\begin{abstract}
Information extraction (IE) aims to produce structured information from an input text, e.g., \textit{Named Entity Recognition} and \textit{Relation Extraction}. Various attempts have been proposed for IE via feature engineering or deep learning. However, most of them fail to associate the complex relationships inherent in the task itself, which has proven to be especially crucial. For example, the relation between 2 entities is highly dependent on their entity types. These dependencies can be regarded as complex constraints that can be efficiently expressed as logical rules. To combine such logic reasoning capabilities with learning capabilities of deep neural networks, we propose to integrate logical knowledge in the form of first-order logic into a deep learning system, which can be trained jointly in an end-to-end manner. The integrated framework is able to enhance neural outputs with knowledge regularization via logic rules, and at the same time update the weights of logic rules to comply with the characteristics of the training data. We demonstrate the effectiveness and generalization of the proposed model on multiple IE tasks.
\end{abstract}

\section{Introduction}
Information extraction (IE) involves the identification of important information from a piece of input text and is a fundamental step towards knowledge inference. Various problems can be categorized as IE tasks, e.g., \textit{Named Entity Recognition (NER)}, \textit{Entity Linking}, \textit{Opinion Target Extraction (OTE)}, \textit{Relation Extraction (RE)}, etc. In this work, we target at 2 challenging IE tasks including OTE and end-to-end RE. Given an input text, end-to-end RE aims to extract target entities as well as entity relations~\cite{liji14}. For example, given the sentence ``\textit{Rome is in Lazio province and Naples in Campania}'', the task requires the identification of \textit{Rome}, \textit{Lazio}, \textit{Naples} and \textit{Campania} as \textit{location} entities, and the relation between \textit{Rome} and \textit{Lazio} as \textit{Located\_In}, same for the relation between \textit{Naples} and \textit{Campania}. The task of OTE aims to identify opinion targets within an opinionated text~\cite{Hu04}, e.g., \textit{service staff} in ``\textit{The service staff in this restaurant is very kind}''.

Deep neural networks (DNNs) have been widely used for various IE tasks. Existing works adopted convolutional neural networks~\cite{xu18,adel17} and recurrent/recursive neural networks~\cite{Wang16,miwa14} to learn context-aware and high-level features to facilitate predictions. Pointer networks have also been proposed for relation extraction~\cite{katiyar17}. Despite their advantage over low-level feature engineering, the complex networks make learning harder when the amount of training data is insufficient, which is the case for many IE tasks. Moreover, the automation in DNNs makes it challenging to inject prior knowledge to guide the training process. On the opposite, symbolic logic systems provide an effective way to express complex domain knowledge in terms of logic rules and have proven to be advantageous when data is scarce. Inspired by the cognitive process that learns from both experiences and background knowledge, recent years have witnessed a growing interest in combining deep learning with logic reasoning~\cite{Manhaeve18,dong2018} mostly for solving logical problems.

To enhance the extraction performance in the NLP domain, we propose to incorporate domain knowledge as logic rules that are integrated into the representation learning system through a unified framework. The proposed model consists of a deep learning module as well as a logic module, where the deep learning module contains a transformer-style neural network to learn a rich feature representation for each word. The transformer model computes complex word-level correlations in multiple dimensions regardless of context distance~\cite{vaswani17}, and has shown promising results in several NLP tasks, e.g., semantic role labeling~\cite{zhixing18}. We believe this mechanism could be more beneficial to propagate information between related entities, compared to other deep models. The multi-head attention weight indicate the interactions between each pair of words which can be further fed into a relation classifier. The logic module is composed of a set of logic rules represented by First-order Logic (FOL). These rules explicitly specify the complex relationships in the output label space, which could not be handled using simple constraints. For example, a FOL rule, $Live\_In(Z, X)\wedge person(Z) \Rightarrow location(X)$, specifies that if the relation between two entities is \textit{Live\_In} and the first entity is of type \textit{person}, then the second entity should have type \textit{location}.

To associate distributed features with logic reasoning, we integrate the deep learning module and the logic module through 2 operations: 1) We design some mapping functions such that the information from the neurons could be passed to the logic system. Specifically, the neural outputs are treated as the inputs to the logic module, which combined with probabilistic logic operators, produces the logic outputs. Hence the outputs from the logic module reflects both neural learning and logic interactions among correlated atoms. Furthermore, a learnable weight is assigned to each logic rule to indicate its confidence level. The learnable weight for each rule makes the logic system more flexible and adaptable to specific training dataset, where a higher weight makes the corresponding rule more important within the corpus. 2) A discrepancy loss is proposed to measure the disagreement between the deep learning module and the logic system, which is minimized to allow for regularization of DNNs via logical knowledge. The discrepancy loss prompts the update of neural parameters towards rule-constrained directions, and at the same time adjusts the rule weights to be compatible with specific corpus.

To summarize, the proposed framework has the following contributions: 1) We use transformer mechanism for IE tasks to fully exploit interactions among the \textbf{input space}, which is also indicative for relation predictions. 2) We use logic rules to enforce complicated correlations in the \textbf{output space} and integrate these rules into the distributed representation learning system with a joint learning mechanism to achieve joint inference. To the best of our knowledge, this is the first work for information extraction that unifies DNN with logic knowledge in a rather \textit{smooth} way to benefit learning of each other. 3) We introduce a general framework for knowledge fusion through discrepancy minimization, which can be adopted in various DNN models. We also demonstrate its effectiveness on different IE tasks.

\section{Related Work}
\noindent \textbf{Information Extraction}
Various approaches have been proposed for entity and relation extraction, either through a pipeline procedure, or a joint inference framework. The pipeline strategy first learns an entity extraction model and then independently predicts relations based on the extracted entities~\cite{chan11,lin16}. This strategy suffers from error propagation. 
To solve this problem, joint inference is proposed to learn shared information between two subtasks by sharing parameters~\cite{miwa16,katiyar17,Bekoulis18,Bekoulis18b} and adopting novel tagging scheme to further model task interactions~\cite{roth04,liji14,miwa14,gupta16,zheng-etal-2017-joint,zhang17,adel17,sun18,shaolei18,Dai19}.

For opinion target extraction, existing works either used pre-defined rules/features to identify the targets~\cite{Hu04,Qiu11,Li10}, or applied deep learning models with sequence labeling strategy~\cite{Yin16,Wang16,Wang17,lixin17,xu18}. However, DNNs only implicitly exploit the input interactions, without controlling what is learned. \citeauthor{Yu19}~\shortcite{Yu19} used integer linear programming with explicit constraints for joint inference as a post-processing step.

\noindent \textbf{Deep Learning with Logic Rules}
The combination of neural learning systems with symbolic rules has long since been proposed, known as neural-symbolic systems~\cite{Garcez12,Manhaeve18,dong2018,Sourek18} that construct a network or connect the distributed systems with given rules for reasoning and inference in logic domains. \citeauthor{xu18semanticloss} treated logic knowledge as semantic regularizsation in the loss function. 
The injection of logic rules in NLP tasks was recently proposed in~\cite{rocktaschel15,guo16} for relation and knowledge graph learning that embed logic into the same space as distributed features in a single system.
\citeauthor{hu16}~\shortcite{hu16} fused logical knowledge into deep models through posterior regularization. Logic rules were also used as evidences to construct adversarial sets~\cite{minervini17,minervini-riedel-2018}, or as a form of indirect supervision~\cite{hai2018} to improve model training. \citeauthor{li2019}~\shortcite{li2019} augmented deep learning models with logic neurons.
In this work, we propose to combine DNN with logic in a smooth way, which adopts probabilistic logic instead of 0/1 hard assignments to facilitate backpropagation through the whole framework. Instead of inserting logic within the DNN architecture, we use a discrepancy loss to progressively bridge the gap between deep learning and logic rules.

\section{Problem Definition \& Motivation}
We use end-to-end RE, which aims to jointly extract entities and their relations, as a motivating task to describe our proposed method. Denote by $\mathcal{E}$ and $\mathcal{R}$ the set of possible entity types and relation categories, respectively.\footnote{Note that the task of OTE can be regarded as a special case of RE, where there is only one entity type.} Given an input sentence $\{w_1, w_2, ..., w_m\}$, entity extraction involves both entity segmentation as well as entity typing. We use BIO encoding scheme combined with entity types to form the sequence of output labels $y \!=\! \{y_1, y_2, ..., y_m\}$, where $y_i \!\in\! \{\textup{B-E}_j, \textup{I-E}_j, \textup{O}\}_{E_j \!\in \mathcal{E}}$. For example, B-PER (I-PER) indicates the beginning (inside) position of an entity of type \textit{person}. 
Relation extraction aims to output a set of triplets ($e_1, e_2, r$), where $e_1$ and $e_2$ represents the first and second entity, respectively, and $r \!\in\! \mathcal{R}$ indicates the relation type between them. In this work, we treat entity extraction as a sequence labeling problem and relation extraction as a classification problem based on the identified entities.

\begin{figure}
    \centering
    \includegraphics[width=0.9\columnwidth]{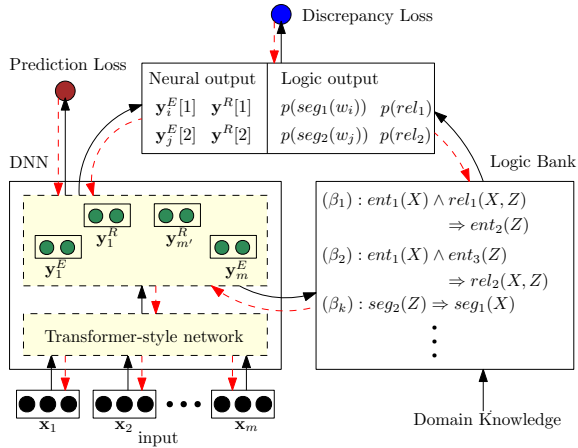}
    \caption{The overall architecture.}
    \label{fig:overview}
\end{figure}

A key motivation behind our proposed method is the complex correlation inherent both in the input and output spaces. In the input space, there exist intensive interactions among entities within a sentence to facilitate information propagation. For example, if \textit{Rome} is extracted as a \textit{location} entity and has close relationship with \textit{Lazio}, it may help to identify \textit{Lazio} as another entity. To exploit these interactions, we use the transformer mechanism with multi-head self-attentions to generate a correlation factor for each pair of words, which is injected for both entity and relation predictions. However, DNNs can only implicitly capture some correlations without actually enforcing specific relationships.  For example, if we know that one of the entities is of type \textit{person} and the relations between the entities is \textit{Live\_In}, the other entity should be of type \textit{location}. Even within entity predictions, there also exists dependencies enforcing possible positions within an entity, e.g., label ``O'' cannot be followed by ``I''. These relationships are especially crucial for the task, and yet few existing works have taken them into account. Although sequence labeling models, like Conditional Random Fields (CRFs)~\cite{crf01icml}, are able to encode certain segmentation rules implicitly, they may not learn the optimal strategy when there is insufficient training data. Furthermore, they fail to model more complex relational rules. A few works treat the complex relationships among objects as constraints within the objective function. However, it is non-trivial to express those complex constraints, and the resultant optimization is challenging. Moreover, the constraints cannot be updated to align with the training corpus. In the literature, symbolic logic rules are well-known to be effective at modeling complex semantic relationships. For example, a dependency between entity types and relations can be expressed using FOL as $person(X) \wedge Live\_In(X, Z) \!\Rightarrow\! location(Z)$. When the training data is insufficient, as is the case for IE, logic rules provide crucial clues to assist learning.

Compared with the pipeline approach which is prone to error propagation, joint inference on related tasks, e.g., entity recognition and relation classification, has shown to be effective for (end-to-end) information extraction. Though joint learning has been proposed in deep architectures, most existing models fail to explicitly enforce the consistency among separate tasks. To address this problem, it is desirable to integrate logic rules specifying task relationships for joint inference. At this point, we propose to unify deep learning with logic rules in an end-to-end learning framework.
To align discrete symbolic system with distributed representation learning, we propose to compute logic rules in a probabilistic way and define mapping functions to map the continuous output from a DNN to the logic units. Furthermore, a discrepancy loss is proposed that measure the discrepancy between the DNN outputs and the logic outputs to make these two modules consistent with each other. The discrepancy loss is able to regularize the DNN through domain knowledge, and at the same time update the logic module to comply with the training data.

\section{Methodology}
The overall architecture of the proposed method is shown in Figure~\ref{fig:overview}. It consists of 3 components, namely a deep neural network, a logic bank and a discrepancy unit. The DNN component takes a sequence of words as the input and finally produces a prediction vector for each word (and possibly candidate relations). The logic bank is fed with general domain knowledge that is easy to obtain and formalizes the knowledge as a set of first-order logic rules. Note that we assign a non-negative weight to each logic rule to indicate its confidence level which is updated according to the training corpus. The output from DNN is fed into the logic module to produce logic output. The discrepancy unit is responsible for aligning neural outputs with outputs from the logic bank. Specifically, we compute a distance between the output distributions from DNN and logic module with the aim to minimize the distance throughout the learning process.

\subsection{First-Order Logic}\label{sec:fol}
FOL is formed from \textit{constants}, \textit{variables} and \textit{predicates} with propositional connectives $\wedge$, $\vee$, $\neg$ and quantifiers. To avoid confusion, we use upper-case letters ($A$) to represent variables and lower-case letters ($a$) to represent constants. An \textit{atom} is an $n$-ary predicate with $n$ arguments ($R(A,B)$). A ground atom assigns constants to all of its arguments ($R(a,b)$). 
A clause can be written in the form of a rule: $B_1 \wedge ... \wedge B_k \Rightarrow H$, where $H$ is called the consequent of the rule and $B_1\wedge ... \wedge B_k$ is the precondition. 
The \textit{grounding} of a clause is a substitution that maps each occuring variable in the clause to a constant: $B_{1(\phi)}\wedge ... \wedge B_{n(\phi)} \Rightarrow H_{(\phi)}$, where $\phi$ denotes a substitution. 
A \textit{Herbrand interpretation} is a mapping that assigns a truth value to each ground atom. To make it a \textit{Herbrand model}, all the logic formulas should be satisfied. To find a Herbrand model, a feasible method is through the \textit{immediate consequence operator}, which is a mapping $T_p$ from Herbrand interpretation to itself:
\begin{eqnarray}\label{eq:consequence}
    T_p(I) = \{H_{(\phi)}|&\!\!\!\!\!\!\!\!\!&(B_1\wedge ...\wedge B_n \Rightarrow H)\in \mathcal{P},\nonumber\\ &\!\!\!\!\!\!\!\!\!&\{B_{1(\phi)},...,B_{n(\phi)}\}\in I\},
\end{eqnarray}
where $I$ is a Herbrand interpretation, $\mathcal{P}$ is a set of clauses. Given known grounded atoms, we can find other grounded atoms as immediate consequences of the logic formulas. In our formulation, we use neural networks to simulate the immediate consequence operator and applies probabilistic logic where each formula is assigned a confidence score and each grounded atom has a continuous truth value within $[0,1]$ to indicate its probability of being true.

\subsection{Deep Learning Module}
\begin{figure}
    \centering
    \includegraphics[width=0.7\columnwidth]{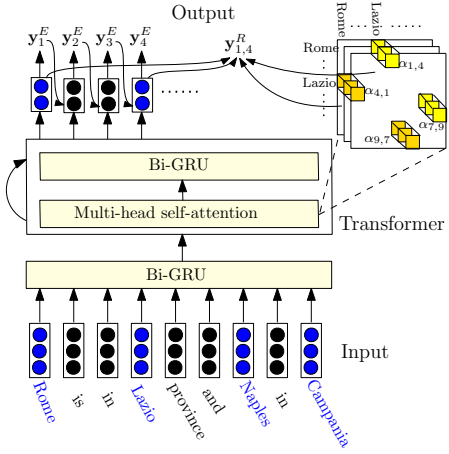}
    \caption{Deep learning module with transformer.}
    \label{fig:transformer}
\end{figure}

The deep learning component is modeled as a transformer-style network consisting of multiple layers of self-attentions and Bi-GRU in order to model both sequential and distant dependencies. A concrete structure is shown in Figure~\ref{fig:transformer}. Specifically, given a sequence of input words, we first construct an input embedding $\mathbf{x}_i=[\mathbf{x}^e_i:\mathbf{x}^p_i]$ for each word by concatenating its pre-trained word embedding $\mathbf{x}^e_i$ and its associated POS-tag embedding $\mathbf{x}^p_i$. To incorporate sequential context interactions, a Bi-GRU model with parameters $\Theta$ is firstly applied on top of $\mathbf{x}_i$ to generate hidden representations $\mathbf{h}_i$ by considering both forward and backward information. Mathematically, we denote this process by
\begin{eqnarray}
  \mathbf{h}_i
  \!=\! [\overrightarrow{\mathbf{h}}_i:\overleftarrow{\mathbf{h}}_i]
  \!=\! [f(\mathbf{x}_i, \overrightarrow{\mathbf{h}}_{i-1}; \Theta):f(\mathbf{x}_i, \overleftarrow{\mathbf{h}}_{i+1}; \Theta)],\nonumber
\end{eqnarray}
where $\overrightarrow{\mathbf{h}}_i$ and $\overleftarrow{\mathbf{h}}_i$ indicate the forward and backward GRU output, respectively.

Subsequently, $\mathbf{h}_i$ is further fed into a transformer model consisting of multiple layers, where each layer stacks a Bi-GRU network on top of a multi-head self-attention module. Specifically, the self-attention network transforms its input $\mathbf{h}^t_i$ at $t$-th layer to $\tilde{\mathbf{h}}^t_i$ via a series of attention mechanism. For ease of illustration, we drop the superscript $t$ in the sequel. Mathematically, given a query vector corresponding to an input vector $\mathbf{h}_i$, each attention head computes one type of interactions between itself and other positions within the sequence and produces a transformed hidden representation:
\begin{eqnarray}
    \tilde{\mathbf{h}}^c_i &=& \sum_{j=1}^m \alpha^c_{ij} (\mathbf{W}^c_v \mathbf{h}_j), \mbox{ and} \\
    \boldsymbol{\alpha}^c_{i} &=& \textup{softmax} (\frac{(\mathbf{W}^c_q \mathbf{h}_i)(\mathbf{W}^c_k \mathbf{H})}{\sqrt{d}}),
\end{eqnarray}
where $\mathbf{H}$ is a matrix consisting of $\mathbf{h}_i$ as column vectors, and $d$ is the dimension of $\mathbf{h}_i$. $\{\mathbf{W}^c_v, \mathbf{W}^c_q, \mathbf{W}^c_k\}$ are transformation matrices corresponding to the $c$-th attention head. We define $C$ different transformations for multi-head mechanism, where each transformation accounts for one representative interaction space.
For IE, each head is regarded as computing a different relation between 2 words. A final hidden vector is produced via
    $\tilde{\mathbf{h}}_i \!=\! \mathbf{W}[\tilde{\mathbf{h}}^1_i:...:\tilde{\mathbf{h}}^C_i]$.

Denote the final feature representation after applying Bi-GRU in the last transformer layer $T$ as $\mathbf{h}^T$. The neural outputs for entity prediction $\mathbf{y}^E$ are generated through a fully-connected layer followed by a softmax layer:
\begin{eqnarray}
\mathbf{s}^E_i &=& \tanh(\mathbf{W}^E_h [\mathbf{h}^T_i:\mathbf{x}^l_{i-1}] + \mathbf{b}^E_h), \\
p(\mathbf{y}^E_i|\mathbf{x}) &=& \textup{softmax}(\mathbf{W}^E_y \mathbf{s}^E_i + \mathbf{b}^E_y),\label{eqn:yE}
\end{eqnarray}
where $\mathbf{x}^l_{i-1}$ denotes the entity label embedding of the previous token. The injection of label embeddings implicitly informs entity label dependencies. Similarly, the relation prediction $\mathbf{y}^R$ is produced for each pair of extracted entities via
\begin{eqnarray}
\mathbf{s}^R_{i,j} \!\!\!\!&=&\!\!\!\! \tanh(\mathbf{W}^R_h [\mathbf{v}_i\!:\!\mathbf{v}_j\!:\!\boldsymbol{\alpha}_{i,j}\!:\!\boldsymbol{\alpha}_{j,i}] + \mathbf{b}^R_h), \\
p(\mathbf{y}^R_{i,j}|\mathbf{x}, \mathcal{E}) \!\!\!\!&=&\!\!\!\! \textup{softmax}(\mathbf{W}^R_y \mathbf{s}^R_{i,j} + \mathbf{b}^R_y),\label{eqn:yR}
\end{eqnarray}
where $\mathbf{v}_i=[\mathbf{h}^T_i:\mathbf{x}^l_i]$, $\mathcal{E}$ indicates the set of extracted entities, and $\boldsymbol{\alpha}_{i,j}$ represents the multi-head attention weight vector between $w_i$ and $w_j$. The attention vectors contain explicit correlation information that could assist relation prediction, as will be verified later in experiments.

\subsection{Logic Fusion via Discrepancy Loss}
As discussed, the DNN module introduced in the previous section is able to exploit word-level interactions implicitly via feature learning, but fails to consider more complex relationships among outputs. For example, the relationship between entity types and their relations, which are non-trivial to be injected into DNNs, but can be effectively formulated as logical rules. We construct 2 types of FOLs to specify the relationships among entity and relation labels. The first type focuses on the dependencies of segmentation labels, denoted by $seg_b(Z) \!\!\Rightarrow\!\! seg_a(X)$, which means if the segmentation label for variable $Z$ is $b$, the segmentation label for variable $X$ is deduced as $a$, with $a, b\!\!\in\!\!\{\textup{B, I, O}\}$. An example is $seg_B(w_i) \!\!\Rightarrow\!\! seg_O(w_{i-1})$ that enforces the previous word to have label O when the current word is the beginning of an entity. The second rule models the correlations between entity types and relations, e.g., $entity_c(X) \wedge rel_l(X, Z) \Rightarrow entity_d(Z)$, which means relation $l$ only appears when the first and second entity has type $c$ and $d$, respectively.

To make logic rules compatible with DNN outputs, we adopt probabilistic logic where each atom is assigned a continuous value in $[0,1]$ and convert Boolean operations to work on probabilistic units. Hence, we define the following: 
\theoremstyle{definition}
\begin{definition}
A mapping $\Gamma$ from the language of FOL $\mathcal{L}$ to the continuous space $\mathbb{R}$ ($\Gamma: \mathcal{L} \rightarrow \mathbb{R}$) is defined as follows,
\begin{itemize}
    \item $\Gamma(P) \!=\! \mathbf{y}(P)$ with $P$ representing a grounded atom and $\mathbf{y}$ as the neural output. For example, $\Gamma(entity_c(w_i)) = \mathbf{y}^E_i[c]$, where $c$ indicates the index of the entity type. This can be interpreted as: the probability for $w_i$ belonging to $entity_c$ equals to its corresponding neural output $\mathbf{y}^E_i[c]$.
    \item $\Gamma(P_1\wedge ... \wedge P_n)=\sigma(a_0(\sum_{i=1}^n \Gamma(P_i) - n) + b_0)$.
    \item $\Gamma(P_1\vee ... \vee P_n)=\sigma(a_1\sum_{i=1}^n \Gamma(P_i) + b_1)$.
    \item $\Gamma(\lnot P)=1-\Gamma(P)$.
\end{itemize}
\end{definition}
\noindent Here $\sigma$ indicates the sigmoid activation function. The last 3 mappings are able to approximate the semantics of logic operators according to~\cite{Sourek18}. With these mapping functions, we compute a soft-version of the \textit{immediate consequence operator} using (\ref{eq:consequence}) to produce the output $Y_{\mathcal{L}}$ from the logic module. 
Specifically, the value of the consequent atom $H_{(\phi)}$ in each rule is deduced by applying $\Gamma$ on the rule body $B_{(\phi)}$, given a grounded clause $B_{(\phi)} \Rightarrow H_{(\phi)}$, where $B_{(\phi)}$ denotes the conjunction $B_{1(\phi)}\wedge ...\wedge B_{n(\phi)}$.
\begin{eqnarray}
    Y_{\mathcal{L}}(H_{(\phi)}) &=& \Gamma (B_{1(\phi)}\wedge ...\wedge B_{n(\phi)}) \nonumber \\
    &=& \sigma(a_0(\sum_{i=1}^n \Gamma(B_{i(\phi)}) - n) + b_0) \nonumber \\
    &=& \sigma(a_0(\sum_{i=1}^n \mathbf{y}(B_{i(\phi)}) - n) + b_0).\label{eqn:logic}
\end{eqnarray}
A detailed procedure to produce logic output is shown in Algorithm~\ref{alg}. Given neural outputs $\{\mathbf{y}^E_i\}_{i=1}^m$ in \eqref{eqn:yE} and $\{\mathbf{y}^R_l\}_{l=1}^{m'}$ in \eqref{eqn:yR} for each sentence, and a set of rules $\{B^{(k)} \!\Rightarrow\! H^{(k)}\}_{k=1}^K$, the logic system produces a logic output $u^E_{i,k}$ for each word and $u^R_{l,k}$ for each relation corresponding to each rule $r_k$ as following: For each rule $B^{(k)} \!\Rightarrow\! H^{(k)}$, we find its satisfying groundings $\phi$ for $B^{(k)}$ from neural predictions and generate the logic output $Y_{\mathcal{L}}(H^{(k)}_{(\phi)})$ for its consequent atom $H^{(k)}_{(\phi)}$ using (\ref{eqn:logic}). For example, given the rule $entity_c(X) \wedge rel_l(X, Z) \Rightarrow entity_d(Z)$, if the neural model predicts $w_i$ as $entity_c$ and the relation between $w_i$ and $w_j$ as $rel_l$, $(X=w_i, Z=w_j)$ is regarded as a satisfying grounding. Then the logic output for the consequent atom $entity_d(w_j)$ will be produced as $\sigma(a_0(\mathbf{y}^E_i[c]+\mathbf{y}^R_{i,j}[l]-2)+b_0)$. We use $\Phi^k$ and $\Gamma^k$ to collect the deduced consequent groundings and their logic values, respectively. The final logic output for each word $u^E_{i,k}$ (relation $u^R_{l,k}$) for each rule $r_k$ is obtained by aggregating the logic values across all the situations when acting as a consequent atom.

\begin{algorithm}[t!]
    \caption{Deep Logic}\label{alg}
    \footnotesize
    \begin{algorithmic}
    \STATE {\bf Input:} Neural softmax outputs $\{\mathbf{y}^E_i\}_{i=1}^m$ (entities) and $\{\mathbf{y}^R_l\}_{l=1}^{m'}$ (relations) for each sentence.
    \STATE {\bf Output:} $\{u^E_{i,k}\}_{i=1,k=1}^{m,K}$ (entities) and $\{u^R_{l,k}\}_{l=1,k=1}^{m',K}$ (relations)
    \STATE {\bf Initialize:} $u^E_{i,k}=0$, $u^R_{l,k}=0$ for $i\in\{1,...,m\}, l\in\{1,...,m'\}, k\in\{1,...,K\}$.
    \STATE Collect feasible groundings of rule head and rule body.
    \FOR {each rule $r_k: B^{(k)}\Rightarrow H^{(k)}$}
    \STATE 1: Initialize $\Phi^k=\{\}$, $\Gamma^k=\{\}$
    \STATE 2: Find a grounding $\phi$ such that each $B^{(k)}_{j(\phi)}$ in $B^{(k)}_{(\phi)}$ is true according to neural predictions $\{\mathbf{y}^E_i\}_{i=1}^m$, $\{\mathbf{y}^R_l\}_{l=1}^{m'}$
    \STATE 3: Update $\Phi^k \leftarrow \Phi^k \cup \{H_{(\phi)}^{(k)}\}$, $\Gamma^k \leftarrow \Gamma^k \cup \{Y_{\mathcal{L}}(H^{(k)}_{(\phi)})\}$
    \ENDFOR
    \STATE Compute logic output for rule heads
    \FOR {$k$ from 1 to $K$}
    \STATE Initialize $c^E_i=0$ for $i\in\{1,...,m\}$, $c^R_l=0$ for $l\in\{1,...,m'\}$
    \FOR {$(\phi,\gamma)\in(\Phi^k,\Gamma^k)$}
        \STATE Return the exact word $w_i$ or relation $r_l$ that corresponds to grounding $\phi$
        \STATE Update $u^E_{i,k}\!\leftarrow\! u^E_{i,k}\!+\!\gamma$, $c^E_i\!\!\leftarrow\! c^E_i\!+\!1$ or $u^R_{l,k}\!\leftarrow\! u^R_{l,k}\!+\!\gamma$, $c^R_l\!\!\leftarrow\! c^R_l\!+\!1$
        \ENDFOR
    \STATE $u^E_{i,k}\leftarrow u^E_{i,k} / c^E_i$, $u^R_{l,k}\leftarrow u^R_{l,k} / c^R_l$
    \ENDFOR

    \end{algorithmic}
\end{algorithm}

By making the DNN and the logic module compatible with each other, we can measure their discrepancy $\ell_{D}(\mathcal{F}, \mathcal{L})$ by comparing the distributions of their outputs:
\begin{eqnarray}
\!\!\!\!\!\!\!\!\!\!\!\!\lefteqn{\ell_{D}(\mathcal{F}, \mathcal{L}) = \mathbb{E}_{x\sim \mathcal{X}}(d(Y_{\mathcal{F}}(x), Y_{\mathcal{L}}(x)))}, \nonumber \\
&& = \frac{1}{K}\!\!\sum_{\{B^{(k)} \Rightarrow H^{(k)} \}} \!\!\frac{1}{|\Phi^k|}\!\! \sum_{\phi\in \Phi^k} \!\!\beta_k d(\mathbf{y}(\phi), \mathbf{u}_k(\phi)), \label{eqn:disloss}
\end{eqnarray}
where $Y_{\mathcal{F}}(x)$ and $Y_{\mathcal{L}}(x)$ denote the neural output and logic output, respectively. $\Phi^k$ collects the consequent atoms whose precondition is satisfied. We further assign a confidence weight $\beta_k \!\in\! [0,1]$ for each rule to indicate its confidence and adjust its contribution to the discrepancy loss. The higher the weight, the more penalty to be given when neural outputs disagree with logic outputs. We use Mean-Squared-Error as the distance metric $d(\cdot,\cdot)$, because it provides a better gradient flow for a more stable training process.

\subsection{Training}
The integrated model can be trained end-to-end with gradient descent by minimizing $\ell = \ell_{Y} + \ell_{D}$, where $\ell_{Y}$ is the prediction loss for the deep learning model. Here we use cross-entropy loss for both entity and relation predictions:
\begin{eqnarray}
    \ell_{Y} \!=\! -\frac{1}{N} \!\!\sum_{n=1}^{N} (\log p (\hat{\mathbf{y}}^E_n | \mathbf{x}_n) \!+\! \log p (\hat{\mathbf{y}}^R_n | \mathbf{x}_n, \mathcal{E}_n)),
\end{eqnarray}
where $p (\hat{\mathbf{y}}^E_n | \mathbf{x}_n)=\prod_{i=1}^{|s_n|}p (\mathbf{y}^E_i=\hat{\mathbf{y}}^E_i | \mathbf{x}_n)$ using (\ref{eqn:yE}). $|s_n|$ indicates the length of the $n$th sentence. Similar procedure applies to $p (\hat{\mathbf{y}}^R_n | \mathbf{x}_n, \mathcal{E}_n)$ using (\ref{eqn:yR}). $\mathcal{E}_n$ denotes the set of extracted entities for the $n$-th sequence. The backpropagation procedure is revealed in Figure~\ref{fig:overview} via (dashed) downward arrows. Specifically, the discrepancy loss updates both neural and logic outputs, together with rule weights according to (\ref{eqn:disloss}) through gradient descent. To restrict the logic weights within $[0,1]$, we apply a sigmoid function such that $\beta_k=\sigma(\beta'_k)$. Then the gradient of the logic weight becomes
\begin{equation}
    \frac{\partial \ell_D}{\partial \beta'_k} = \frac{1}{K|\Phi^k|}\sum_{\phi\in\Phi^k}d \sigma(\beta'_k)(1-\sigma(\beta'_k)),
\end{equation}
where $d=d(\mathbf{y}(\phi), \mathbf{u}_k(\phi))$. The gradients for logic output $\mathbf{u}_k(\phi)$ is further passed back to neural logits $\mathbf{y}(B_{(\phi)})$ according to (\ref{eqn:logic}), which combined with the classification loss, updates all the parameters within the neural model. Ideally, the discrepancy loss will punish the situation when the neural output highly differs from the logic output. In this case, the deep model will modify its network to be more aligned with the logic rules. On the other hand, the logic module will adapt its weights as well as the mappings that are passed back to the neurons. For example, if the deep module predicts \textit{Rome} as \textit{location}, \textit{Lazio} as \textit{location} and their relation as \textit{OrgBased\_In} (which is wrong). When feeding them into the rule $loc(Rome) \wedge OrgBased\_In(Lazio, Rome) \Rightarrow org(Lazio)$, the logic output for $entity_{org}(Lazio)$ will be high, different from the neural output. In this case, the discrepancy revises its rule body to decrease the incorrect neural output for relation \textit{OrgBased\_In}.

\section{Experiment}
To demonstrate the effectiveness of our proposed method, we conduct experiments on 5 datasets from 2 tasks: 
\\ \noindent \textbf{OTE}: We use Restaurant and Laptop reviews from SemEval 2014 and 2016~\cite{sem14,Sem16}. \\
\noindent \textbf{End-to-End RE}: 1) \textbf{TREC}: entity and relation dataset introduced in~\cite{roth04}. It has 4 entity types: \textit{others}, \textit{person}, \textit{location} and \textit{organization}, and 5 relations: \textit{Located\_In}, \textit{Live\_In}, \textit{OrgBased\_In}, \textit{Work\_For} and \textit{Kill}. We follow the preprocessing from~\cite{gupta16} 2) \textbf{ACE05}: annotated data with 7 coarse-grained entity types and 6 coarse-grained relation types between entities. We follow the same setting as~\cite{liji14}.

For the OTE task, we follow the setting in~\cite{Wang16} by first pre-training the word embedding using \textit{word2vec}~\cite{Mik13} on Yelp Challenge dataset\footnote{http://www.yelp.com/dataset\_challenge} and electronic dataset in Amazon reviews\footnote{http://jmcauley.ucsd.edu/data/amazon/links.html} for restaurant domain and laptop domain, respectively. For RE task, the word embedding is pre-trained on wikipedia corpus using \textit{Glove}~\cite{glove14}. For all experiments, the dimension for word embedding and POS embedding is set to 300 and 50, respectively. The hidden layers has dimension 200. We set label embedding with dimension 25. Following~\cite{vaswani17}, we also use positional encoding that is added to the input vectors. The multi-head self-attentions adopts 10 heads that leads to 10-dim attention weight vectors. 
For RE task, we use scheduled sampling, similar to~\cite{miwa16}. To train the model, adadelta is adopted with initial rate as 1.0 and with dropout rate 0.1. For evaluation, we use micro-F1 scores on non-negative classes. An entity is counted as correct based on exact match. A relation is correct if both of its entities are correct and the relation type matches the ground-truth label. 

\subsection{Results \& Analysis}
\begin{table}[t]
    \centering
    \footnotesize
    \begin{adjustbox}{max width=0.9\columnwidth}
    \begin{tabular}{l|c|c|c}
    \hline
    \hline
     &   Restaurant14 &  Laptop14  & Restaurant16 \\
     \hline
    \cite{Wang16}  &  84.25   &  77.26  &  69.74 \\
    \hline
    \cite{Wang17}  &  84.38  &  76.45  &  73.87  \\
    \hline
    \cite{lixin17}  &  -  &  77.58  &  73.44  \\
    \hline
    \cite{xu18}  &  84.24  &  81.59  &  74.37  \\
    \hline
    \cite{Yu19}  &  84.50  &  78.69  &  -  \\
    \hline
    TransF  &  84.64  &  81.76   &  73.56   \\
    \hline
    Ours  &  \textbf{85.62} &  \textbf{82.46} &  \textbf{74.67} \\
    \hline
    \end{tabular}
    \end{adjustbox}
    \caption{Comparison with baselines on OTE.}
    \label{tab:aspect}
\end{table}
\noindent \textbf{Comparison on OTE task}: Table~\ref{tab:aspect} shows the comparison results for opinion target extraction with popular baselines. The last two rows indicate our proposed models, where \textit{TransF} is the deep learning module without logic integration. Since the OTE task can be viewed as single-class entity extraction, the proposed model can be adapted to this task by ignoring relation predictions. From the results, we can observe that even without logic rules, the transformer model is able to achieve 2-out-of-3 best results compared to existing works.
This proves the effectiveness of transformer for implicit interaction modeling. When considering logic knowledge, we use segmentation rules that enforces possible segmentation labels for 2 adjacent words.
Furthermore, we also incorporate implicit relational rules that state $entity(X) \wedge entity(Z) \Rightarrow related(X,Z)$. This is achieved by keeping the relation prediction layer in the deep learning module. We find this strategy slightly improves our results which will be shown later. The results also show that the integration of logic rules is more effective than CRF by comparing with~\cite{Wang16}. Although \citeauthor{Yu19}~\shortcite{Yu19} incorporated explicit constraints through integer linear programming, the separation from DNN during learning makes it suboptimal. This demonstrates the advantage of our unified framework that associates logic reasoning with representation learning. Clearly, our model achieves the state-of-the-art results on all 3 datasets.

\begin{table}
\begin{center}
\begin{adjustbox}{max width=\columnwidth}
\begin{tabular}{c|l|l|l|l}
\hline
\hline
Setting                             & Model  & Evaluation & Entity & Relation \\ \hline
\multirow{10}{*}{w/ Boundary}     &   (Gupta et al. 2016)     & relaxed    & 92.4   & 69.9     \\ \cline{2-5}     &
\cite{Bekoulis18} & relaxed    & 93.3   & 67.0     \\ \cline{2-5}   &
\cite{Bekoulis18b} & relaxed    & 93.0   & 68.0     \\ \cline{2-5}   &
\cite{miwa14} & strict    & 92.3   & 71.0          \\ \cline{2-5}   &
\cite{adel17} & strict    & 92.1   & 65.3          \\ \cline{2-5}   &
Pipeline       & strict     & 94.1   & 69.7          \\ \cline{2-5}   &
Pipeline+feat       & strict     & 94.5   & 70.2          \\ \cline{2-5}   &
TransF       & strict     & 94.6   & 72.8          \\ \cline{2-5}   &
Ours (w/o) POS       & strict     & 94.3   & 71.7          \\ \cline{2-5}   &
Ours        & strict     & \textbf{95.1}   & \textbf{74.1}          \\ \hline \hline
\multirow{9}{*}{w/o Boundary} &     \cite{miwa14} & relaxed   & 80.7   & 61.0          \\ \cline{2-5}   &
\cite{adel17} & relaxed   & 82.1   & 62.5          \\ \cline{2-5}   &
\cite{Bekoulis18} & strict & 83.0   & 61.0          \\ \cline{2-5}   &
\cite{Bekoulis18b} & strict & 83.6  & 62.0          \\ \cline{2-5}   &
Pipeline        &  strict   &  84.2  & 57.2          \\ \cline{2-5}   &
Pipeline+feat        &  strict   &  84.2  & 58.4          \\ \cline{2-5}   &
TransF        &  strict   &  85.8  & 62.7          \\ \cline{2-5}   &
Ours (w/o) POS        &  strict   &  85.0  & 63.1          \\ \cline{2-5}   &
Ours          & strict    &  \textbf{87.1}  & \textbf{64.6} \\
\hline
\end{tabular}
\end{adjustbox}
\caption{Results of End-to-End RE on TREC.}\label{tab:TREC}
\end{center}
\end{table}

\begin{table*}[t]
    \centering
    \begin{adjustbox}{max width=\textwidth}
    \begin{tabular}{l|c|c|c|c|c|c|c|c|c|c|c}
    \hline
    \hline
         &  (Li and Ji 14)  &  (Miwa and Bansal 16)  &  (Zhang et al. 17)  &  (Sun et al. 18) & (Sun et al. 18)$_G$ & Pipeline & Pipeline+feat & TransF & TransF$_G$ &  Ours & Ours$_G$  \\
    \hline
    Entity  &  80.8  &  83.4  &  83.5  &  83.4 & \textbf{83.6} & 83.2 & 83.2 &  83.4 & 83.5 & 83.4 &  \textbf{83.6} \\
    \hline
    Relation  &  49.5  &  55.6  &  57.5  &  57.8 & \textbf{59.6} & 57.4 & 55.3 &  59.1  &  59.1  & 59.3  & 59.4  \\
    \hline

    \end{tabular}
    \end{adjustbox}
    \caption{Comparison with baselines on relation extraction using ACE2005 dataset.}
    \label{tab:ace}
\end{table*}

\begin{table}[t]
    \centering
    \begin{adjustbox}{max width=\columnwidth}
    \scriptsize
    \begin{tabular}{p{0.6cm}|p{1.6cm}| p{0.5cm}|p{0.5cm}|p{0.5cm}|p{0.5cm}|p{0.7cm}|p{0.5cm}|p{0.7cm}}
    \hline
    \hline
    \multirow{2}{*}{Settings} & \multirow{2}{*}{Models} &  Res14  &  Res16  &  Lap14  &  \multicolumn{2}{c|}{TREC}  & \multicolumn{2}{c}{ACE05}  \\
    \cline{3-9}
    & & \multicolumn{3}{c|}{OTE} & Entity  &  Relation & Entity  &  Relation  \\
    \hline
    \multirow{3}{*}{Entity} & TransF$-\mathbf{x}^l$  &  84.1  &  72.7  &  77.6  &  83.3  &  - & 82.9 & - \\
    \cline{2-9} & TransF  &  84.6  &  73.6  &  81.8  &  84.2  &  -  & 83.2 & - \\
    \cline{2-9} & TransF$+$SR  &  85.0  &  73.9  &  82.1  &  84.9  &  - & 83.2 &  - \\
    \hline
    \multirow{3}{*}{Joint} & TransF$-\boldsymbol{\alpha}$ & - &  - &  - & 84.3  &  60.6  &  83.3  &  57.4 \\
    \cline{2-9} & TransF & - &  - & -  &  85.8  &  62.7 & \textbf{83.4} & 59.1 \\
    \cline{2-9} & TransF$+$SR$+$RR  &  \textbf{85.6}  &  \textbf{74.7}  &  \textbf{82.5}  &  \textbf{87.1}  &  \textbf{64.6}  &  \textbf{83.4} & \textbf{59.3} \\
    \hline

    \end{tabular}
    \end{adjustbox}
    \caption{Comparisons on different model settings.}
    \label{tab:setting}
\end{table}

\noindent \textbf{Comparison on RE task}: The comparison results on TREC is shown in Table~\ref{tab:TREC}. Existing works for this domain consists of 2 different settings and evaluations. The first setting is introduced in~\cite{roth04} that assumes the entity boundaries are given and the task is to predict the entity types and relations. The second setting requires both segmentation and entity typing. Evaluations include relaxed version where an entity is regarded as correct if at least one of its consisting words have the correct type prediction. The strict version only treats a predicted entity as correct given a complete match. When boundaries are given, our model could be easily modified to treat each entity as a single unit for type predictions. To show the effect of joint inference with logic rules, we construct a pipeline model by first predicting entities followed by relation predictions given fixed entity parameters. Another model (Pipeline+feat) further appends rule-based features as 1-hot vectors for relation prediction. Obviously, the pipeline model achieves inferior performance and simple features are far less expressive than logic rules. For both settings, our model achieves the best results with a large margin. Furthermore, we also test our model without the POS embedding, shown as ``Ours (w/o) POS'', which still demonstrates some performance gain.  

Table~\ref{tab:ace} shows the comparison results on ACE05 dataset. Note that \citeauthor{sun18}~\shortcite{sun18} used a non-decomposable global loss with normal local loss 
to jointly train entity and relation extraction. We denote by (Sun et al. 18)
and (Sun et al. 18)$_G$ the model without and with the global loss, respectively. We also implemented the global loss as a fine-tuning step in our method, denoted by the subscript $G$. It is observed that \textit{TransF} achieves the best result for relation extraction and comparable result for entity extraction without sophisticated global training strategy. Global loss does not provide clear benefit for our model. This might indicate that our model already explores global interactions. (Sun et al. 18)$_G$ requires substantial pretraining processes (larger than 500 epochs in our implementation) using the normal loss. As a comparison, our model is more convenient to implement and provides faster convergence, as will be shown later. The logic rules only show slight improvement. We deduce the reason to be the characteristics of the data. Indeed, there are very few absolute relationships between entity types and relations. In most cases, two related entity types can have multiple relation categories, making explicit rules less useful. 

\noindent \textbf{Comparison for different model settings}: To demonstrate the effect of each component, we conduct experiments on different model settings, with the statistics shown in Table~\ref{tab:setting}. The objective is to analyze the effect of interactions within entities themselves, as well as the correlations between entities and relations. We first remove the relation prediction component to examine the results for solely entity extraction, denoted as `Entity'. We construct 3 different settings to compare the results among standard transformer model, removing label embedding (TransF$-\mathbf{x}^l$), and integrating only segmentation rules (TransF$+$SR) with discrepancy loss. As shown, the label embedding slightly improves all the performances which proves its ability to implicitly exploit the segmentation dependencies. However, more improvements are observed by explicitly incorporating logic rules. When jointly predicting entities and relations, we could observe slight improvements upon entity extraction using transformer. This shows positive correlations between entities and their relations. For this joint task, we examine the effect of distributed features (TransF$-\mathbf{\alpha}$) as well as symbolic rules (TransF$+$SR$+$RR) on the final results. By removing attention weight features, TransF$-\mathbf{\alpha}$ achieves inferior results on relation predictions, demonstrating the advantage of attentions for relation modeling. More significant improvement could be observed when incorporating the relationships between entities and relations as explicit rules.

\begin{figure}
    \centering
    \includegraphics[width=0.8\columnwidth]{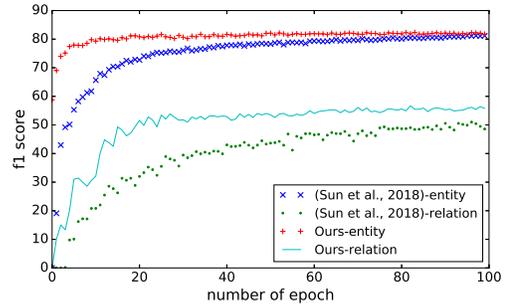}
    \caption{Performance trend comparison on 2 models.}
    \label{fig:perf_trend}
\end{figure}

\noindent \textbf{Convergence Comparison}: (Sun et al. 18)$_G$ requires a great amount of pre-training epochs using the normal local loss (at least 500 epochs) for the base model to reach reported performance. As a comparison, our model with the normal loss converges much faster, as demonstrated in Figure~\ref{fig:perf_trend} on the test data. Compared with~\cite{sun18}, our model reaches to a relatively high performance within 20 epochs. The result for entity extraction has faster convergence rate compared to relation prediction but stays in a rather stable state, which is comparable to~\cite{sun18} in subsequent epochs. The relation prediction performance for our model substantially increases in the first place, while still keeps improving later on. We can observe that within 100 epochs, our model achieves clear performance gain over~\cite{sun18} on relation extraction.

\section{Conclusion}
In this work, we propose a novel unified model to associate distributed learning with symbolic rules. The integrated framework is able to pass information from the neural model to the logic module and compute a discrepancy loss between these two components, which is minimized to update the whole network. The marriage between these two systems could regularize deep learning in the form of knowledge distillation. On the other hand, the logic system is also updated in terms of rule weights to adapt to specific data domain. Experimental results demonstrate the advantage of combining DNNs and logic for joint inference.

\section{Acknowledgements}
This work is supported by NTU Nanyang Assistant Professorship (NAP) grant M4081532.020, Singapore MOE AcRF Tier-2 Grant MOE2016-T2-2-060, and the Lee Kuan Yew Postdoctoral Fellowship program, Singapore.

\small
\bibliography{aaai20}
\bibliographystyle{aaai}

\end{document}